# Transfer Learning for OCRopus Model Training on Early Printed Books

Christian Reul[1], Christoph Wick[1], Uwe Springmann[2], and Frank Puppe[1]

[1] *Chair for Artificial Intelligence and Applied Informatics*
[2] *Kallimachos Center for Digital Humanities*
*University of Würzburg, Germany*
*Email: <firstname.lastname>@uni-wuerzburg.de*

**Abstract**

A method is presented that significantly reduces the character error rates for OCR text obtained from OCRopus models trained on early printed books when only small amounts of diplomatic transcriptions are available. This is achieved by building from already existing models during training instead of starting from scratch. To overcome the discrepancies between the set of characters of the pretrained model and the additional ground truth the OCRopus code is adapted to allow for alphabet expansion or reduction. The character set is now capable of flexibly adding and deleting characters from the pretrained alphabet when an existing model is loaded. For our experiments we use a self-trained mixed model on early Latin prints and the two standard OCRopus models on modern English and German Fraktur texts. The evaluation on seven early printed books showed that training from the Latin mixed model reduces the average amount of errors by 43% and 26%, respectively compared to training from scratch with 60 and 150 lines of ground truth, respectively. Furthermore, it is shown that even building from mixed models trained on data unrelated to the newly added training and test data can lead to significantly improved recognition results.

## 1 Introduction

Starting from Breuel et al.'s 2013 groundbreaking paper [1] the application of recurrent neural networks with LSTM architecture to the field of OCR of historical printings has made excellent progress [2] [3] [4], although it was previously considered nearly impossible for the case of incunabula[1] [5]. Character error rates (CERs) below 5% are now routinely possible for even the earliest printings. However, this can only be achieved by training specific recognition models for each individual book, or at least for books coming from the same print shop and printed with the same font. This does not scale up very well for conversion of the already available substantial amount of scanned book pages from the 15th to 18th century [6]. Ideally one would construct models resembling the so-called *polyfont* or *omnifont* recognition models employed by standard OCR engines such as Tesseract or ABBYY Finereader. They achieve very good overall recognition rates to more recent printings from the 19th century onwards, often with CERs of 1% and below.

The prime factor preventing the construction of effective models for earlier printings is the scarcity of ground truth (GT) training material, i.e. diplomatic[2] transcriptions of real printings. The production of GT is a costly and slow manual process, which in the case of early printings often entails specialized knowledge to decode the meaning of palaeographic glyphs into Unicode characters.

---

[1] Incunabula are the first modern printings from the period 1450-1500.

[2] A diplomatic transcription is one that records only the characters as they appear on the support, with minimal or no editorial intervention or interpretation.

This barrier can be overcome for modern printings by the creation of synthetic training material, starting from available electronic text, which gets rendered into synthetic images using available computer fonts, often with some noise added to make the model more robust.[3] For early printings we lack the pertinent fonts containing the specific shapes and glyphs used by individual printing shops. In the incunabula period from 1450-1500, as many as 2,000 individual print shops employing 6,000 different fonts have been identified and collected in printed tables accompanying Haebler's monumental *Typenrepertorium der Wiegendrucke*.[4] Furthermore, a recognition model for early printings does not just depend on specific fonts but also on the interword distance, as printers meticulously cared for justified right margins and ran words closely together to make this happen if no convenient break point was possible. The difficulty of getting tokens correctly recognized becomes apparent when trained individual models have wrongly split or merged words as their most frequent error. The next more frequent error types are insertions, deletions and substitutions such as $e \leftrightarrow c$.

Because the synthetic method currently does not work for early printings and real GT is both scarce and expensive to produce, we have to look for other means to build workable models.

As modern and historical font shapes (glyphs) are not totally different, a simple idea is to reuse the models trained on modern fonts and use them as a starting point for continued training on some historical GT. Thus, one could hope to reach a certain level of CER with less historical GT than if we trained a model from scratch. In the following we explore this idea and report some experiments that show if and to what extent this expectation is justified.

A note on terminology: The alphabet on which a recurrent neural network is trained is also called *codec*, as the alphabet is internally represented by numbers to which the alphabet gets encoded and which at the end gets again decoded to alphabet characters. A pure or individual model is trained on a single book (which might contain different typesets, e.g. upright and cursive) and is contrasted with a *mixed model* trained on GT relating to different books, which mostly also means different typographies (different fonts, different interword distances). Models trained on synthetic material in different languages are also called mixed models by us even when all languages are represented by the Latin script using Antiqua fonts, as there are specific national typographic idiocracies leading to different codecs (e.g. the usage of accents in French texts, or different punctuation marks).

Chapter 2 describes related work, chapter 3 gives details of the pretrained models and their respective GT used as well as our modifications of the OCRopus code, chapter 4 relates our experiments and their outcomes which are then discussed in chapter 5. At the end follows chapter 6 with the conclusions and ideas for future work.

## 2   Related Work

Breuel et al. [1] used their own open source tool OCRopus[5] to recognize modern English text and German Fraktur from the 19th century by training mixed models, i.e. models trained on a variety of fonts, typesets, and interword distances from different books. The English model was trained on 95,338 text lines from the UW-III dataset[6] consisting of modern English prints. Applying the model to 1,020 previously unseen lines from the same dataset yielded a character error rate (CER) of 0.6%.

---

[3] The new Tesseract neural network models for Latin scripts have been constructed using synthetic images with 4500 fonts: https://github.com/tesseract-ocr/tesseract/wiki/TrainingTesseract-4.00

[4] http://tw.staatsbibliothek-berlin.de/

[5] https://github.com/tmbdev/ocropy

[6] http://isis-data.science.uva.nl/events/dlia//datasets/uwash3.html

The training set for the Fraktur model mostly consisted of around 20,000 mostly synthetically generated text lines. The resulting model was evaluated on two books of different scan qualities yielding CERs of 0.15% and 1.37%, respectively.

An approach not only mixing different types but also various languages was promoted by Ul-Hasan and Breuel in [7]. They generated synthetic data for English, German and French and used it for training language specific models as well as a mixed one. As expected, the language specific models performed best when applied to test data of the same language yielding CER of 0.5% (English), 0.85% (German) and 1.1% (French). However, recognizing a mixed set of text data with the mixed models also led to a very low CER of 1.1%. Despite being carried out exclusively on synthetic data this experiment indicates a certain robustness of OCRopus regarding varying languages in mixed models.

The idea of training mixed models was adapted to early prints by Springmann et al. in different application scenarios. In [2] their corpus consisted of twelve books printed with Antiqua types between 1471 and 1686 with a focus (ten out of twelve) on early works produced before 1600. It was divided into two distinct sets of six books and a mixed model was trained on both of them. Evaluating each model on the respective held-out books mostly yielded CERs of under 10% (with two exceptions). Obviously, these results are far off the numbers reported above which can be explained due to the vastly increased variety of the types. Still, the trained models provide a valid starting point for further model improvements through individual training.

During a case study on the RIDGES corpus[7], a similar experiment was conducted on 20 German books printed between 1487 and 1870. Again, by training mixed models on half of the books and evaluating on the held-out data impressive recognition results of around 5% CER in average were achieved. As expected, the individually trained models performed even better, reaching an average CER of around 2%.

While to the best of our knowledge there is no suitable related work regarding transfer learning in the field of OCR, it was applied successfully to a variety of other tasks. Yosinski et al. performed experiments on the transferability of features in deep neural networks [8]. They used the ImageNet dataset[8], which at the time of the described experiments consisted of close to 1.3 million labeled training images and 50,000 test images, with each image labeled with one of 1,000 classes. After randomly splitting the classes in half they first performed a pretraining on one half before training and finally testing on the remainder. This approach yielded lower error rates compared to the default method, i. e. only training and testing on data with fitting classes. So even after an extensive period of fine-tuning on fitting data, the features learned during the first steps still lingered and led to notably improved recognition accuracies.

Wick and Puppe applied the same method in [9] using even more diverse data sets. In order to assign the correct species to images of leafs they first performed a pretraining on the Caltech-256 dataset[9], consisting of over 30,000 images assigned to 256 classes like animals, tools, vehicles or fictional characters. Afterwards, they built from the obtained network by training on real leaf images. Despite the diversity of the two sets of training data, the pretraining showed a significant positive effect on the classification accuracy.

Obviously, these examples of transfer learning used far deeper networks than OCRopus with only a single hidden layer, resulting in a dramatically increased number of parameters and consequently, more opportunities to learn and maintain useful low-level features. Nonetheless, we still expect a noteworthy impact of pretraining, since scripts in general should be expected to show a higher degree of similarity than e.g. oak leafs and Homer Simpson.

---

[7] http://korpling.org/ridges presented in [3]

[8] http://www.image-net.org/

[9] http://www.vision.caltech.edu/Image_Datasets/Caltech256/

## 3 Materials and Methods

We first introduce our evaluation corpus consisting of books we partially transcribed to support various projects. We expect our approach to work best with models trained on data as similar as possible to these books. Therefore, we use available data from our evaluation corpus to train a historical mixed model for OCRopus. In addition, we use two less similar mixed models trained on newer types.

Furthermore, some necessary changes regarding the OCRopus code are described, which enable us to extend and reduce the set of characters available to a model in a flexible way.

### 3.1 Books

The experiments were performed on seven early printed books (see Table 1).

Table 1: Books used for Evaluation.

| ID/Year | Language | GT Train | GT Test |
|---------|----------|----------|---------|
| 1476 | German | 1,000 | 2,000 |
| 1488 | German | 1,500 | 2,678 |
| 1495 | German | 1,000 | 1,114 |
| 1500 | Dutch | 1,250 | 1,250 |
| 1505 | Latin | 1,500 | 1,789 |
| 1509 | Latin | 1,500 | 1,500 |
| 1572 | Latin | 791 | 750 |

To avoid unwanted side effects only lines from running text parts were used and headings, marginalia, page numbers, etc. were excluded. Figure 1 shows some example lines.

Figure 1. Different example lines from the seven books used for evaluation. From top to bottom: excerpts from books 1476, 1488, 1495, 1500, 1505, 1509, and 1572.

The books 1495, 1500, 1505 and 1509 are editions of the Ship of Fools and were digitized as part of an effort to support the Narragonien digital project at the University of Würzburg[10]. Despite their similar content these books differ considerably from an OCR point of view since they have been printed in different print shops using varying typefaces and languages (Latin, German and Dutch). 1488 was gathered during a case study of highly automated layout analysis [4]. 1476 is part of the Early New High German Reference Corpus[11] and 1572 was digitized in order to be added to the AL-Corpus[12]. All books above the horizontal line in Table 1 are printed in broken scripts (Fraktur in the wider sense), the rest used Antiqua types.

---

[10] http://kallimachos.de/kallimachos/index.php/Narragonien

### 3.2 Mixed Models

Our first model was trained on historical books printed in Latin using the same data as in [2] (abbreviated LH for Latin Historical). After training on 8,684 lines for 109,000 iterations the best model was chosen by evaluating all resulting models on 2,432 previously unseen test lines. The lowest achieved CER was 2.92% after 98,000 training steps.

Additionally, we used the freely available OCRopus standard models for English (ENG)[13] and German Fraktur (FRK)[14] introduced in [1] and described above.

### 3.3 Utilizing Arbitrary Pretrained Models in OCRopus

OCRopus in its original form already allows to load existing models and continue training from there. However, the default functionality only covers the case where a training is stopped (deliberately or not) and restarted using the exact same alphabet. While this suffices to ensure that the training process doesn't get lost, it cannot be applied to material with additional characters.. Therefore, the following adjustments on code level had to be made. The corresponding source code is available at Github[15].

#### 3.3.1 Extending the Codec

While mixed models are usually trained on a variety of different books and therefore comprise a rather comprehensive alphabet it still is likely for them to sooner or later encounter previously unknown characters. For any (mixed) model it is impossible to recognize these glyphs it has never seen during training, so these glyphs constitute *blind spots* for the recognition process. Even worse, if a character isn't part of the codec it can never be learned. Therefore, the model must be able to grow.

Figure 2 illustrates the extension and reduction (see next section) of the codec. The bidirectional LSTM based network used by OCRopus consists of two layers. A single LSTM layer processes each pixel wide slice of the text line, thus its input dimension equals the line height in pixel. The number of time steps $T$ equates the line length. The LSTM layer produces a vector $h$ for each time step. Its size $H$ remains fixed for each model and is given by the number of states in the single hidden layer of the OCRopus network (default = 100). The last layer represents a matrix multiplication where the weight matrix $M$ is multiplied with the current $h$ producing an output $o$ for each character in the codec. Each character is represented in $M$ by a vector of size $H$, containing the weights determined during the training process. Consequently, the dimensions of $M$ are $H$ times $C$, with $C$ being the codec size. The predictions with probability $P(c)$ for each character $c$ in the codec is generated by applying a softmax function to $o$. Since each single character in the codec is given by a row in $M$, a codec extension can be achieved by adding additional rows to $M$. For each attached row, an additional entry in the output is appended. Yet, the application of the softmax function ensures that the output remains a valid probability distribution $P(c)$. The new weights in $M$ are initialized randomly and have to be trained to produce meaningful results.

---

[11] http://www.ruhr-uni-bochum.de/wegera/ref/index.htm

[12] http://arabic-latin-corpus.philosophie.uni-wuerzburg.de

[13] http://www.tmbdev.net/en-default.pyrnn.gz

[14] http://tmbdev.net/ocropy/fraktur.pyrnn.gz

[15] https://github.com/ChWick/ocropy/tree/codec_resize

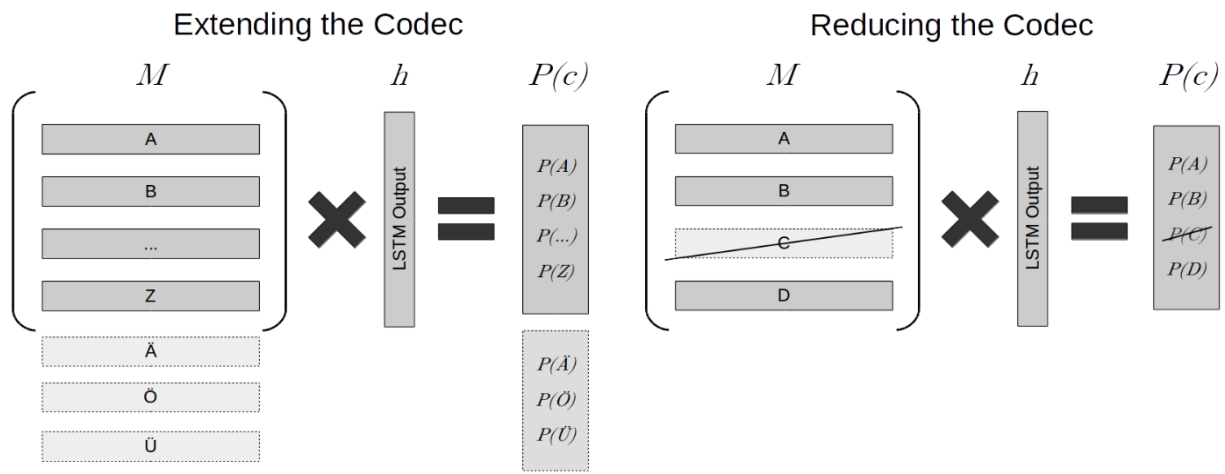

Figure 2. Schematic view of extensions (left) or reductions (right) of the output matrix of the network whose rows correspond to the codec.

### 3.3.2 Reducing the Codec

The just described problem regarding characters missing from the codec could obviously be bypassed by simply bloating the codec. However, this is impractical for two reasons. First, the bigger the codec the slower the training and recognition process becomes. Second, when refining a mixed model towards an individual one for a single book the goal is to minimize the number of recognizable characters without risking blind spots. Obviously, a large codec also makes misrecognitions more likely, especially if it contains several very similar characters. For example, LH contains several *e* characters with various diacritica on top, e.g. *éèêë*, which are customarily employed in early printings. However, in books that do not contain these diacritics they only add potential for confusions.

The right sketch of Figure 2 shows the process of removing single characters from the output matrix $M$. By deleting a complete row the corresponding output probabilities $P(c)$ is removed, too. Retraining the network is not necessary since the softmax ensures that the output still is a valid probability distribution.

## 3.4 Defining a Whitelist Containing Immune Characters

Especially when working with small amounts of GT it is likely that these transcribed lines don't comprise all characters that occur throughout the entire book. In this case applying the approach described above will lead to blind spots. Therefore, we implemented a whitelist (WL) containing characters that won't be removed from the codec even if they don't occur in the GT used for training: a-z, A-Z, 0-9.

## 4 Experiments

In order to examine our hypothesis that building from an existing model holds clear advantages compared to training from scratch ('default model') we conducted several experiments whose outcomes are reported in this section. After explaining the general methodology of our training and evaluation procedure, we conduct the first experiment comparing the default approach with a training starting from the LH model. Next, since suitable models in terms of printing type, age and language are often not available we test the OCRopus standard models ENG and FRK and still hope for improvements compared to the default training. Furthermore, we expect the gains of our pretraining approach to correlate with the number of lines used for training. Since more lines lead to stronger models, the room for improvement gets smaller and we therefore await smaller gains. In our final experiment we replace the mixed LH model by a model trained on a single similar book with the expectation to achieve even bigger improvements.

### 4.1 Setup and Methodology

The initial setup consisted of two main steps. First, for each book about half of the available GT was set aside for evaluation. The remaining individual GT was split up in five different training/test sets on which models were trained and their results averaged to reduce the impact of variance. To ensure maximal comparability, both, the initial training/test/evaluation split as well as the individual training sets were kept fixed for all experiments.

The actual model training using OCRopus was always carried out for a fixed number of iterations until no further notable improvements were observed. An amount of 10% to 15% of the training lines were set aside before training to act as a test set in order to determine the best model, i.e. the one that produced the lowest CER on the test set. Finally, the best models are used to recognize the held out evaluation data to determine the final result.

### 4.2 Building from the Latin Mixed Model

In this first experiment we compare a training starting from the LH model with one starting from scratch. When using the LH model all trainings were performed twice, once with building the codec from the available GT and once with adding the whitelist WL as described in section 3.4. All experiments were carried out for 60 and 150 lines of GT since usually 60 lines are a good starting point and 150 lines represent just enough lines to already train relatively strong individual models (150) without reaching the point of diminishing return. Table 2 shows the results.

Table 2. Resulting CERs when using the raw Latin Hist model (*LH only*), models trained from scratch (*Def*) and by building from the Latin Hist model without (*LH*) and with (*+WL*) utilizing the whitelist. All CERs and improvement rates (*Gain*) given in % for the seven books. The last line shows the average (*AVG*).

| Book | LH only | 60 Lines of GT (52 Train, 8 Test) | | | | | 150 Lines of GT (130 Train, 20 Test) | | | | |
|---|---|---|---|---|---|---|---|---|---|---|---|
| | | Def | LH | | +WL | | Def | LH | | +WL | |
| | CER | CER | CER | Gain | CER | Gain | CER | CER | Gain | CER | Gain |
| 1476 | 31.12 | 8.21 | 5.35 | **35** | 5.17 | **37** | 4.00 | 3.11 | **22** | 3.04 | **24** |
| 1488 | 35.28 | 7.60 | 3.53 | **54** | 3.49 | **54** | 2.88 | 2.22 | **23** | 2.22 | **23** |
| 1495 | 42.79 | 12.67 | 6.26 | **51** | 6.14 | **52** | 5.83 | 4.03 | **31** | 4.04 | **31** |
| 1500 | 37.61 | 5.03 | 3.58 | **29** | 3.42 | **32** | 2.95 | 2.42 | **18** | 2.29 | **22** |
| 1505 | 17.23 | 6.19 | 5.32 | **14** | 4.79 | **23** | 3.70 | 3.43 | **7** | 3.40 | **8** |
| 1509 | 5.05 | 6.31 | 2.85 | **50** | 2.06 | **67** | 2.81 | 2.24 | **20** | 1.44 | **49** |
| 1572 | 10.40 | 2.43 | 1.58 | **35** | 1.61 | **34** | 1.72 | 1.27 | **26** | 1.26 | **27** |
| **AVG** | 25.64 | 6.92 | 4.07 | **38** | 3.81 | **43** | 3.41 | 2.67 | **21** | 2.53 | **26** |

The achieved CERs show that building from a mixed model leads to superior individual models compared to using the available GT by itself. As expected, the improvement rates decrease with more GT for training and increase with adding a whitelist of basic characters. The average gain when utilizing 60 lines of GT is 43%, from a CER of 6.92% without pretraining to 3.81%. This is nearly as good as using a considerable more expensive GT of 150 lines without pretraining, having a CER of 3.41%. With pretraining (including the whitelist), a CER of 2.53% is achieved using 150 lines of GT, with an average gain of still 26% over the default approach. Interestingly, the improvements don't necessarily correlate with the performance of the LH on its own. For example, the book where the LH model did worst on (1495) still experiences one of the highest boosts among all books.

Since adding the whitelist shows a clearly positive effect (average gain of 5%) all remaining experiments were performed by including the whitelist.

The gained accuracy vary considerably. For example, book 1505 shows the least improvement over the default approach (but still 23% and 8%, respectively). Most likely this is caused by the fact that the distances between two characters in book 1505 are considerably smaller compared to all other books used for training and testing (see Figure 1, line 5).

### 4.3 Utilizing the OCRopus Standard Models

The creation of high quality historical mixed models is a cumbersome task and there aren't many publicly available. Therefore, we investigated the effect of pretraining on a mixed model trained on different but easily available data, in this case using the OCRopus standard models ENG and FRK introduced in section 2. Table 3 sums up the results.

Table 3. Resulting CERs from models trained by following the default approach (Def) compared to building from the Latin Hist model (*LH*) and the Standard OCRopus models *ENG* and *FRK*. All CERs and improvement rates given in %.

| Book | 60 Lines of GT (52 Training, 8 Test) | | | | | | |
|---|---|---|---|---|---|---|---|
| | **Def** | **LH** | | **ENG** | | **FRK** | |
| | CER | CER | Gain | CER | Gain | CER | Gain |
| **1476** | 8.21 | 5.17 | 37 | 5.21 | 37 | **4.49** | **45** |
| **1488** | 7.60 | **3.49** | **54** | 4.32 | 43 | 4.12 | 46 |
| **1495** | 12.67 | **6.14** | **52** | 6.89 | 46 | 6.31 | 50 |
| **1500** | 5.03 | **3.42** | **32** | 4.11 | 18 | 3.49 | 31 |
| **1505** | 6.19 | **4.79** | **23** | 5.44 | 12 | 5.09 | 18 |
| **1509** | 6.31 | **2.06** | **67** | 2.94 | 53 | 4.09 | 35 |
| **1572** | 2.43 | **1.61** | **34** | 1.91 | 21 | 2.25 | 8 |
| **AVG** | 6.92 | 3.81 | 43 | 4.40 | 33 | 4.26 | 33 |
| Book | 150 Lines of GT (130 Train, 20 Test) | | | | | | |
| | **Def** | **LH** | | **ENG** | | **FRK** | |
| | CER | CER | Gain | CER | Gain | CER | Gain |
| **1476** | 4.00 | **3.04** | **24** | 3.21 | 20 | 3.12 | 22 |
| **1488** | 2.88 | **2.22** | **23** | 2.68 | 7 | 2.38 | 17 |
| **1495** | 5.83 | 4.04 | 31 | 4.12 | 29 | **3.89** | **33** |
| **1500** | 2.95 | **2.29** | **22** | 2.50 | 15 | 2.47 | 16 |
| **1505** | 3.70 | **3.43** | **7** | 3.45 | 7 | 3.53 | 7 |
| **1509** | 2.81 | **1.44** | **49** | 1.93 | 31 | 2.40 | 15 |
| **1572** | 1.72 | **1.26** | **27** | 1.25 | 27 | 1.57 | 8 |
| **AVG** | 3.41 | 2.53 | 26 | 2.73 | 19 | 2.77 | 17 |

Although the gains of the ENG und FRK models are slightly lower than for the more similar LH model, they are still impressive: 33% on average for training with 60 lines of GT and 19% (ENG) and 17% (FRK), respectively for training with 150 lines of GT compared to the default approach. As expected, ENG outperforms FRK on the books using Antiqua types (books 1509 and 1572), while FRK has higher gains for Fraktur types (books 1476, 1488, 1495, 1500, and 1505).

### 4.4 Varying the Number of Lines

To further test the applicability of our approach, we repeated some of the experiments by varying the amount of GT in five steps from 30 to 60, 100, 150, and 250 lines. For reasons of clarity the results of only three representative books (1476, 1495, and 1572) are displayed in Figure 3. The remaining books showed the same tendencies.

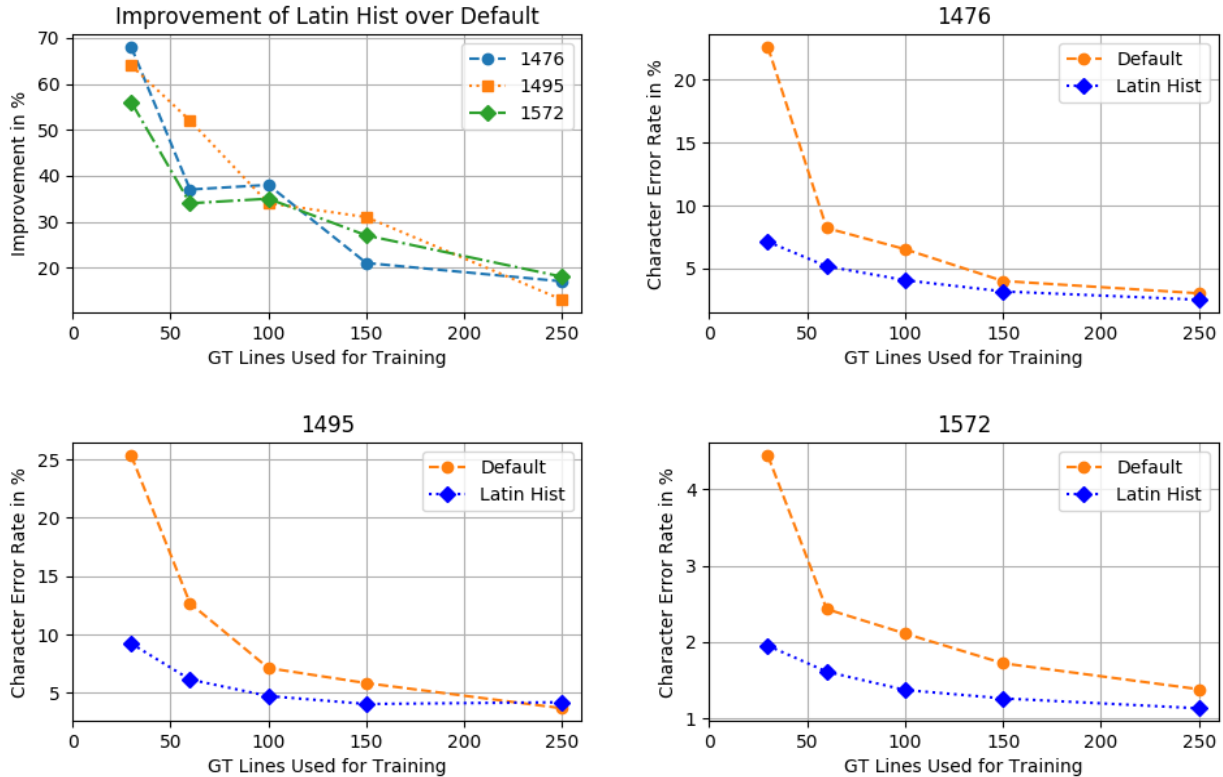

Figure 3. Effects of building from the LH model compared to the default approach for a varying number of lines showing the improvement rates for three different books (top left) and the resulting CERs for 1476 (top right), 1495 (bottom left) and 1572 (bottom right).

As expected and in line with previous experiments, the achievable improvements decrease when increasing the amount of available GT. While for a small amount of lines (30 and 60) the CER is reduced by at least one third and up to two thirds, this effect almost vanishes for most books when approaching 250 lines.

### 4.5 Incorporating Individual Models

Next, we want to examine if building from a model trained on an individual book similar to the new data can yield even better results than the mixed model approach we utilized thus far. We measured similarity by determining the CER obtained by models trained on individual books and by the mixed models LH, ENG, and FRK on the GT data of the new book. For books 1476, 1505, 1509 and 1572 the LH model performed best and they were therefore excluded from further experiments. The 1488 model achieved the lowest CER on 1495 and vice versa and 1500 got recognized best by the individual model of 1476. Consequently, we trained new models for 1488, 1495 and 1500 by building from the models of 1495, 1488 and 1476, respectively. Of course, each individual model was excluded from the pool when processing the book it was trained on. Table 4 shows the obtained results.

Table 4. Resulting CERs from models trained by following the default approach (Def) compared to building from the Latin Hist model (*LH*) and the best fitting individual model. All CERs and improvement rates given in %.

| Book | | 60 Lines of GT (52 Training, 8 Test) | | | | | | |
|------|------|------|------|------|------|------|------|------|
| | | **Mixed Model (LH)** | | | **Best Fitting Individual Model** | | | |
| | Def | Raw | Trained | **Gain** | Model | Raw | Trained | **Gain** |
| **1488** | 7.60 | 34.56 | 3.49 | 54 | 1495 | 15.58 | **3.23** | **58** |
| **1495** | 12.67 | 43.26 | 6.14 | 52 | 1488 | 16.26 | **5.82** | **54** |
| **1500** | 5.03 | 37.23 | **3.42** | **32** | 1476 | 27.67 | 4.58 | 9 |
| Book | | 150 Lines of GT (130 Train, 20 Test) | | | | | | |
| | | **Mixed Model (LH)** | | | **Best Fitting Individual Model** | | | |
| | Def | Raw | Trained | **Gain** | Model | Raw | Trained | **Gain** |
| **1488** | 2.88 | 35.42 | **2.22** | **23** | 1495 | 16.07 | 2.35 | 19 |
| **1495** | 5.83 | 42.95 | 4.04 | 31 | 1488 | 16.49 | **3.58** | **36** |
| **1500** | 2.95 | 37.60 | **2.29** | **22** | 1476 | 27.52 | 2.66 | 10 |

The results do not show a clear tendency: in three cases, pretraining with the mixed LH model showed higher gains, and in the other three cases, pretraining with the best fitting individual model led to better results. Neither approach shows a significant gain over the other. From this experiment we cannot infer that it is worthwhile to incorporate individual models compared to the robust mixed model for pretraining. However, it has to be said that even the best fitting models only achieved CERs of around 16% or even worse. Therefore, higher gains should be expected when building from individual models, which already fit even better to the new data.

## 5  Discussion

Our experiments showed that building from a pretrained model can significantly reduce the obtainable CER compared to starting the training from scratch. The achievable improvement rates decrease with an increasing amount of GT lines available for training. The effect of a whitelist used to prevent blind spots is reduced when adding more lines since the likelihood for missing characters in the training data goes down.

The evidence that even completely unrelated mixed models also lead to considerable improvement indicates that a pretrained model offers much more than an accurate description of the type(s) it was trained on. Despite the shallow structure of the OCRopus network with only one hidden layer the training seems to benefit a lot from low level features that generalize well like general character shapes, different forms and severity of glyph degradation as well as an improved robustness against noise.

Not a single case occured in our experiments where the proposed approach had a noteworthy negative impact on the recognition result. This seems sensible, since the weights of the network are initialized randomly when training from scratch causing the network to be unable to output anything during the beginning of training before slowly learning the most frequent characters like whitespaces, *e* and *a*. It seems that a pretrained model, which might not match the types at all but at the very least is able to distinguish between character and non-character benefits the training process more than a random initialization. Since the additional required effort when building from a model is negligible our results imply that it is sensible to prefer the pretrained approach over training from scratch, especially when only a low to medium amount of GT is available.

# 6   Conclusion and Future Work

A method to significantly improve the CER on early printed books by building from pretrained models instead of training from scratch was proposed. Our experiments showed that adding fresh GT to an existing model outperforms the default training approach even if GT and model differ considerably, in particular if only a small number of transcribed lines is available. Despite our focus on very early prints using Latin script our experiments suggest that the proposed method should work with a wide variety of prints with diverse scripts and languages and different periods.

A very promising task for the future is the combination of the proposed pretraining approach with our voting procedure introduced in [10]. For the voting to be successful the participating models aren't only required to be precise but also diverse. We have shown that a wide variety of mixed models is likely to have a positive effect on the training outcome. This makes the training of individual models by building from completely different mixed models a very attractive option to gather several powerful, yet highly variant models.

Despite the encouraging results achieved with only one hidden layer, transfer learning tends to be most effective when applied to deeper network structures since the higher amount of parameters allows for the transfer of even more well generalizing features. Since we are currently experimenting on replacing the default OCRopus network by (possibly deeper) Tensorflow[16] networks it will be interesting to see if further gains can be expected.

Furthermore, additional models would be very useful for real world application scenarios, since a suitable model to start training from can save hours of transcription effort. This includes several types: mixed models like LH which are created by collecting and combining real life data, as well as synthetically trained mixed models like ENG or FRK, but also book specific models. Obviously, it is also possible to combine several approaches, for example by taking a small subset of the LH data and train a new model building from ENG or FRK. Since creating GT for models is a time consuming task, especially when aiming for a strong mixed model comprising several books, sharing is key. To lead by example we therefore utilized the books printed in Fraktur used in this paper to train a mixed Fraktur model for early printed books and made it available at GitHub[17] together with the strong individual models used for evaluation in section 4.5, and some test data for all books.

With a growing repository of available models, it makes sense to narrow down the selection before testing on the available GT to find the best fitting model. This can be done by taking attributes like age, the printing type (Antiqua or Fraktur) or if applicable specifics like very small inter character distances into consideration. Thus, the gain of building from pretrained models can be further optimized.

---

[16] https://www.tensorflow.org/

[17] https://github.com/chreul/OCR_Testdata_EarlyPrintedBooks